\documentclass{elsarticle}
\usepackage{graphicx} 
\usepackage{amsmath}
\usepackage{hyperref}

\begin{document}

\title{A Neural Network Approach to Multi-radionuclide TDCR Beta Spectroscopy}
\author[first,second]{Li Yi}
\ead{li.yi@sdu.edu.cn}
\author[first,second]{Qian Yang}
\ead{yangqian2020@sdu.edu.cn}
\affiliation[first]{
    organization={Institute of Frontier and Interdisciplinary Science, Shandong University},
    city={Qingdao},
    postcode={266237}, 
    state={Shandong},
    country={China}
}
\affiliation[second]{
    organization={Key Laboratory of Particle Physics and Particle Irradiation, Ministry of Education, Shandong University},
    city={Qingdao},
    postcode={266237}, 
    state={Shandong},
    country={China}
}

\date{July 2025}

\begin{abstract}
Liquid scintillation triple-to-doubly coincident ratio (TDCR) spectroscopy is widely adopted as a standard method for radionuclide quantification because of its inherent advantages such as high precision, self-calibrating capability, and independence from radioactive reference sources. However, multiradionuclide analysis via TDCR faces the challenges of limited automation and reliance on mixture-specific standards, which may not be easily available. Here, we present an Artificial Intelligence (AI) framework that combines numerical spectral simulation and deep learning for standard-free automated analysis. $\beta$ spectra for model training were generated using Geant4 simulations coupled with statistically modeled detector response sampling. A tailored neural network architecture, trained on this dataset covering various nuclei mix ratio and quenching scenarios, enables autonomous resolution of individual radionuclide activities and detecting efficiency through end-to-end learning paradigms. The model delivers consistent high accuracy across tasks: activity proportions (mean absolute error = 0.009), detection efficiencies (mean absolute error =
0.002), and spectral reconstruction (Structural Similarity Index = 0.9998), validating its physical plausibility for quenched $\beta$ spectroscopy. This AI-driven methodology exhibits significant potential for automated safety-compliant multiradionuclide analysis with robust generalization, real-time processing capabilities, and engineering feasibility, particularly in scenarios where reference materials are unavailable or rapid field analysis is required.

\end{abstract}

\maketitle

\section{Introduction}
\label{sec:intro}
$\beta$ emitting have evolved as indispensable tools in multidisciplinary research, including nuclear facility monitoring, environmental surveillance, and biomedical engineering. Unlike $\alpha$ particles, which exhibit discrete monoenergetic peaks and $\gamma$ rays, which produce characteristic spectral lines, $\beta$ decay generates a continuous energy spectrum. This continuity stems from the shared momentum between the electron emitted and the neutrino. The continuity nature of $\beta$ emission poses unique challenges for the spectral analysis of multiradionuclides with overlapping energy distributions between components. The growing need to analyze mixed $\beta$ emitter systems, such as $^3$H-$^{14}$C dual-nuclide mixtures in biomedical studies, multicontaminant assemblages in environmental assays and in radio pharmaceutical development, has intensified demand for advanced spectral de-convolution techniques. However, intrinsic spectral overlap between $\beta$ emitters, combined with detection efficiency variations, raises significant challenges for accurate absolute activity determination with traditional multi-radionuclide analyzer techniques, particularly when sample composition or detection efficiency are unknown.

Liquid scintillation analysis (LSA) is a well-established technique for detecting low-energy beta emitters, such as tritium-3 and carbon-14, due to its inherent advantage of dissolving the sample directly in the scintillation cocktail, ensuring optimal contact between the emitted $\beta$ particles and the scintillator \cite{lAnnunziata2020book,Hou2020book,Hou2018review}. The Triple-to-Double Coincidence Ratio (TDCR) method for LSA $\beta$ emitting measurement emerged around 1980s to address quenching limitations in liquid scintillation counting.  Double-tube coincidences isolate genuine $\beta$ events from noise, while triple-tube  coincidences serve as a normalization parameter. By calculating the ratio of triple-to-double coincidences, the method enables absolute activity determination without external quench curve calibrations. Recognized as a primary standardization technique for activity measurements by the International Bureau of Weights and Measures (BIPM) \cite{COULON2023110945,BRODA2003585}, TDCR systems have now became worldwide commercial available products, for example, PerkinElmer Tri-Carb and Hidex 300SL. While TDCR system achieves remarkable precision for single-nuclide analyses, it faced challenges in distinguishing different radionuclides in mixtures, especially when their beta spectra overlap or when dealing with high levels of quenching. Current strategies, such as the exclusion method (spectral windowing) and inclusion method (coupled-equation solving)\cite{lAnnunziata2020book}, require laborious external quenching calibrations using standardized sources which prohibits wide application for field-deployable systems or resource-limited laboratories. Moreover, energy window selection in those methods is largely empirical introducing systematic uncertainty in resolving spectrally adjacent nuclides, while rigid mathematical frameworks fail to adapt to nonlinear quenching dynamics, often yielding nonphysical solutions or convergence failures.

In recent years, artificial intelligence (AI) and machine learning (ML) techniques have emerged as powerful tools to address these limitations. Kernel-based Gaussian process regression methods have enabled robust background estimation to enhance isotopic identification accuracy for low signal-to-noise ratio gamma spectra \cite{alamaniotis2013kernel,alamaniotis2024XaI}. Deep learning architectures including fully-connected networks (FCN), convolutional neural networks (CNNs) and attention-based models advanced nuclide identification by exploiting hierarchical feature extraction: CNNs treating spectra as 2D images achieved $>$92\% accuracy for multi-radionuclide mixtures \cite{liu2022novel}; channel-attention mechanism learned to assign higher weights to feature maps that encode the photoelectric peak and Compton edge while suppressing those dominated by background and noise\cite{wang2022multiple}; and a hybrid model combining FCN and CNN is reported to largely outperforming existing state-of-the-art models for interpreting gamma spectral data\cite{galib2021gamma}. Beyond identification, AI has proven effective in inverse problems such as detector response unfolding. Neural networks applied to plastic scintillator gamma spectra reduced relative energy unfolding errors to $<$3.8\% \cite{Heshmati2022gamma}.  Efforts have also be made to enhance nuclear AI model interpretability by equipping CNN-based gamma spectrometry classifiers with saliency heatmaps and class activation mapping, allowing human experts to visually confirm that photopeaks instead of artifacts drive nuclide identification and thereby boosting transparency and trust \cite{gomez2021isotope,wang2022XAI}.

Despite these advancements of AI-driven approaches thrive in gamma spectroscopy \cite{zehtabvar2024gammareview,kamuda2020cnnVsfcn,kim2019gamma,buonanno2020gamma,daniel2020gamma}, their application to $\beta$ spectrum analysis is little explored. This is especially true for multi-radionuclide decomposition involving quenching-induced efficiency corrections in liquid scintillation detectors. This gap limits progress in fields reliant on $\beta$ emitters, highlighting the need for novel AI frameworks tailored to $\beta$ spectral complexities. This work explores the integration of Deep Neutral Network (DNNs) with TDCR-based $\beta$-spectrometry, focusing on algorithmic architecture design, with dataset generation via Monte-Carlo simulation combined with statistical modeling, and validation with $^3$H-$^{14}$C mixed systems. 

\section{Methods}
\label{sec:methods}

\subsection{TDCR System Simulation}
\label{subsec:simulation}

The primary beta particle energy spectra of target radionuclides ($^{3}$H and $^{14}$C) were generated using the Geant4 Monte Carlo toolkit (version 11.0) \cite{agostinelli2003geant4,allison2006geant4,allison2016geant4} with decay parameters sourced from the Evaluated Nuclear Structure Data File (ENSDF) \cite{ensdf2013}. A total of \(1 \times 10^6\) decay events were simulated for each nuclide to ensure statistical convergence. Raw beta energy spectra, denoted as \(dN/dE\) (event count per energy interval), were binned in 1 keV increments.

To model the conversion from beta particle energy to scintillation photon signals, we assumed a linear relationship between deposited energy and scintillation photon yield:

\begin{equation}
N_{\text{photon}} = \alpha \cdot E
\label{eq:photon_yield}
\end{equation}
where \(E\) is the beta particle energy (keV), \(\alpha = 10 \, \text{photons/keV}\) is the scintillator light yield (consistent with commercial liquid scintillators such as Ultima Gold$^{\text{TM}}$ , and \(N_{\text{photon}}\) is the total number of scintillation photons produced. The simulation of quenching effect which will be described in Sec.~\ref{subsec:dataset} is basically realized by adjusting the light yield. 

The TDCR system was modeled assuming three identical virtual PMTs corresponding to channels Ch1, Ch2, and Ch3 with symmetric configurations. Photon detection in each channel was governed by Poisson statistics, where the number of detected photons in channel \(i\) (\(N_i\)) is described as:

\begin{equation}
N_i \sim \text{Poisson}(\lambda_i), \quad \lambda_i = f_i \cdot N_{\text{photon}}^{i}
\label{eq:poisson_statistics}
\end{equation}
where \(f_i = 0.3\) is the PMT quantum efficiency and \(\lambda_i\) is the mean count for channel \(i\) which is assumed to be identical for all three PMTs.

The stronger quenching effect in triple-tube coincidence relative to double-tube coincidence enables TDCR method and forms its foundational principle. AI solves for the detection efficiency by leveraging the implied efficiency contained within the comparison of triple-tube and double-tube coincidence spectra.
The statistical distribution for triple-tube coincidence and double-tube coincidence spectra bases on non-zero counts in specific channel combinations.  For triple-tube coincidence and double-tube coincidence spectra, we derived the probability distributions of total counts \( S = N_1 + N_2 + N_3 \) (where \( N_i \) denotes counts in channel \( i \)) using Poisson statistics and conditional probability.

A double-tube coincidence event is defined as an event with \textit{at least two active channels} (i.e., \( (N_1>0 \cap N_2>0) \cup (N_1>0 \cap N_3>0) \cup (N_2>0 \cap N_3>0) \)). The probability distribution of total counts \( S = N \) under this condition is:

\begin{equation}
\begin{aligned}
P(S=N \mid \text{double-tube coincidence}) &= P(S=N) 
\\&\quad - 3P(S=N \mid \text{single channel active}) \\
&\quad - P(S=N \mid \text{all channels inactive}) 
\end{aligned}
\label{eq:2channel_dist}
\end{equation}
Substituting Poisson statistics (\( N_i \sim \text{Poisson}(\lambda) \), with total counts \( S \sim \text{Poisson}(3\lambda) \) for independent channels), this becomes:

\begin{equation}
\begin{aligned}
P(S=N \mid \text{double-tube coincidence}) &= \text{Poisson}(N; 3\lambda) \\
&\quad - 3\text{Poisson}(N; \lambda)\text{Poisson}(0; \lambda)^2 \\
&\quad - \text{Poisson}(0; \lambda)^3 
\end{aligned}
\label{eq:2channel_poisson}
\end{equation}
where \( \text{Poisson}(x; \mu) = \frac{e^{-\mu}\mu^x}{x!} \) is the Poisson probability  function with mean \( \mu \); The first term on the right hand side \( \text{Poisson}(N; 3\lambda) \) describes unconditioned total counts.
The second term subtracts events with only one active channel, scaled by 3 for all single channel combinations. The third term excludes the all-zero event.

A triple-tube coincidence event requires \textit{all three channels to be active} (\( N_1>0 \cap N_2>0 \cap N_3>0 \)). Its probability distribution is:

\begin{equation}
\begin{aligned}
P(S=N \mid \text{triple-tube coincidence}) &= P(S=N) \\
&\quad - 3P(S=N \mid \text{one channel inactive}) \\
&\quad + 3P(S=N \mid \text{two channels inactive}) \\
&\quad - P(S=N \mid \text{all channels inactive})
\end{aligned}
\label{eq:3channel_dist}
\end{equation}
With Poisson statistics, this simplifies to:

\begin{equation}
\begin{split}
P(S=N \mid \text{triple-tube coincidence}) &= \text{Poisson}(N; 3\lambda)\\
&\quad - 3\text{Poisson}(N; 2\lambda)\text{Poisson}(0; \lambda) \\
&\quad + 3\text{Poisson}(N; \lambda)\text{Poisson}(0; \lambda)^2 \\ 
&\quad - \text{Poisson}(0; \lambda)^3 
\end{split}
\label{eq:3channel_poisson}
\end{equation}
where the second term subtracts events with one inactive channel, scaled by 3. The third term corrects for over-subtraction of events with two inactive channels. The fourth term excludes the all-zero event.

Equation \ref{eq:2channel_poisson} and Equation \ref{eq:3channel_poisson} are used to generate the final double-tube and triple-tube coincidence spectra, capturing the statistical behavior of coincident counting of TDCR systems. Further Poisson statistics can be applied to these two  coincidence light equations to simulate first dynode gain effect.


\subsection{Simulation Dataset Preparation}
\label{subsec:dataset}

A simulation dataset was constructed to enable training and validation of nuclide decomposition for beta spectroscopy in liquid scintillation TDCR counting systems. For target nuclide mixtures (e.g.,  \textsuperscript{3}H and \textsuperscript{14}C), activity proportions were randomly sampled with a unit sum. The composite beta spectrum for each mixture sample was generated by linearly combining the individual nuclide spectra simulated using Geant4 according to these sampled proportions. 

Quenching effects were incorporated by randomly sampling a quenching factor within the range (0, 1], which was applied as a multiplicative scaling factor to the light yield coefficient $\alpha$ in Eq. \eqref{eq:photon_yield}. More sophisticated quenching can be added in future study by adding additional nonlinear term to account for Birks ionization effect\cite{Birks1951} . The double-tube and triple-tube coincidence spectra (Q2 and Q3) were computed by applying Eqs. \eqref{eq:2channel_poisson} and \eqref{eq:3channel_poisson} to the composite beta spectra. The final spectra of TDCR are binned in 1024 channels.

A 5\% random noise component was added to the TDCR Q2 and Q3 spectra to simulate radioactive background contributions and electronic noise encountered in real-world measurements. The detection efficiencies for double-tube and triple-tube coincidence, along with the TDCR ratio, were derived from the ratio of the photoelectron spectra to the original beta spectra. 

The final dataset comprised 10,000 samples, partitioned into training (80\%), validation (10\%), and test (10\%) sets to maintain compositional diversity across splits. The training set enabled the model to learn intrinsic spectral patterns, such as quenching-induced shape changes in Q2/Q3 coincidence spectra, while the validation set ensured these patterns generalized to unobserved data. Finally, the test set containing entirely novel nuclide mixtures and quenching conditions validated the model’s performance. Each sample contained  information on nuclide activity proportions, quenching level, detection efficiencies for both coincidence modes, composite double-tube and triple-tube coincidence spectra, and the calculated TDCR value.

\subsection{Model Architecture and Training Protocol}
\label{subsec:model}

\begin{figure}[htb]
    \centering
    \includegraphics[width=0.8\textwidth]{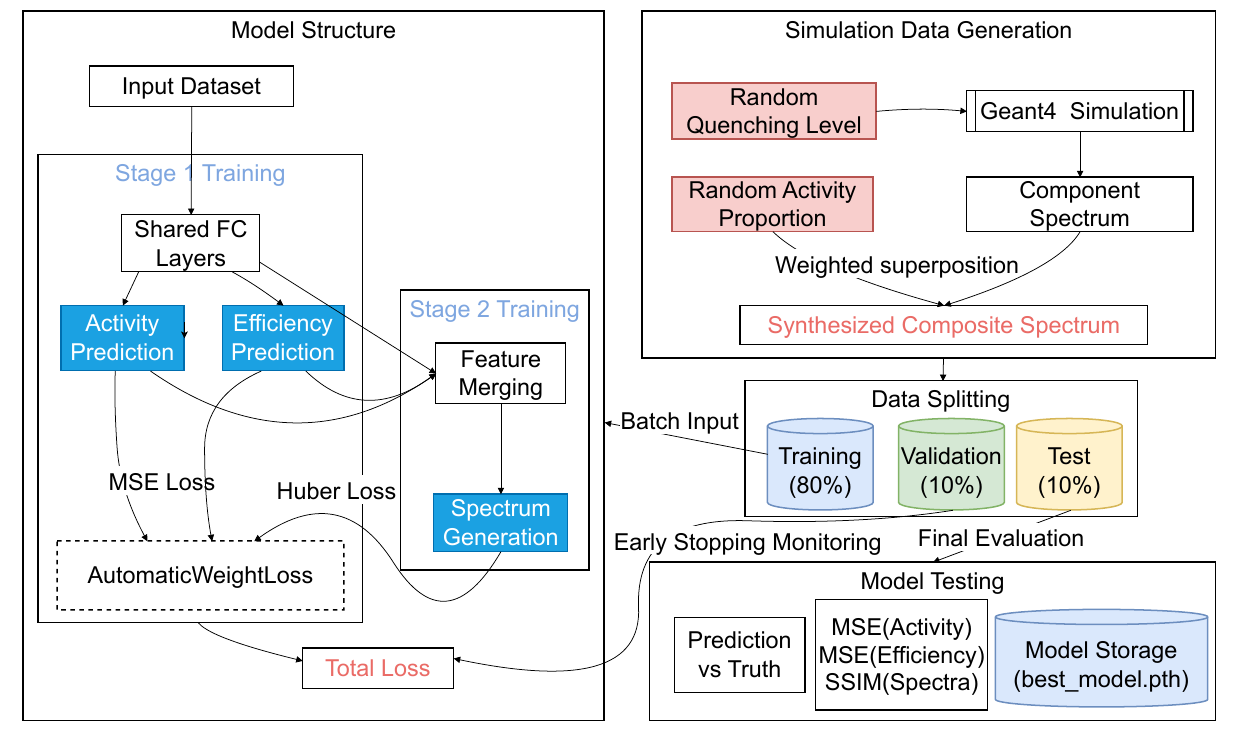}
    \caption{Schematic of the multi-task neural network architecture, showing input processing, shared feature extraction, and task-specific output branches for activity proportion prediction, efficiency estimation, and spectral reconstruction from double-tube and triple-tube coincidence spectra.}
    \label{fig:model_architecture}
\end{figure}

A multi-task neural network was developed to jointly predict nuclide activity proportions and their individual detection efficiencies from the double-tube and triple-tube coincidence spectra. The individual nuclide quenched spectra were also reconstructed. The architecture (Fig. \ref{fig:model_architecture}) consists of three main components:

1. \textbf{Input and Shared Feature Extraction}: The double-tube (Q2) and triple-tube (Q3) coincidence spectra of the mixture is flattened to a one dimensional vector, processed through fully connected layers with ReLU activation \cite{nair2010rectified,glorot2011deep} and batch normalization.  This shared layers learn common spectral features from the correlated nature of spectral information across tasks.

2. \textbf{Task-Specific Branches}:

   - \textbf{Activity Branch}: Receives the shared features and processes them through fully connected layers with dropout regularization \cite{srivastava2014dropout} enabled, followed by an output layer with $n_{\text{nuclides}}$ nodes for the activity proportions.
   
   - \textbf{Efficiency Branch}: Processes shared features through fully connected layer with ReLU activation, outputting $2n_{\text{nuclides}}$ values corresponding to Q2 and Q3 efficiencies for each nuclide. 
   
   - \textbf{Spectrum Branch}: Combines shared features with activity and efficiency predictions into a combined vector, processed through fully connected layers with ReLU activation to reconstruct $2n_{\text{nuclides}} \times 1024$ quenched spectra. 

Training was performed in two stages using the Adam optimizer \cite{Kingma2014}:

- \textbf{Stage 1}: Initial training with the spectrum branch frozen, focusing on activity and efficiency prediction. Loss weights between branches are dynamically adjusted \cite{kendall2018multi} to balance mean squared error (MSE) losses for the two tasks.

- \textbf{Stage 2}: Joint fine-tuning with all branches unfrozen, using a reduced learning rate and extended loss function incorporating Huber loss for spectral reconstruction to reduce sensitivity to outliers.

Early stopping  \cite{prechelt2012early}  was used to prevent overfitting by terminating training when validation loss plateaued. Model performance was evaluated using MSE and mean absolute error (MAE) for numerical outputs, and Euclidean distance and structural similarity index (SSIM) \cite{wang2004image}  for spectral reconstructions. Post-processing applied non-negativity constraints and proportion normalization to ensure physical plausibility of predictions.

\section{Results and Discussion}
\label{sec:result}

The multi-task neural network demonstrated robust performance across nuclide activity prediction, detection efficiency estimation, and spectral reconstruction tasks. For activity proportion prediction, the model achieved a MSE of $0.0003$ and MAE of $0.009$, indicating high precision in quantifying nuclide proportions.  Minor degradation occurred for mixtures with extreme nuclide ratio imbalance ($\leq5\%$ $^3$H or $^{14}$C in $^3$H/$^{14}$C mixtures). This limitation arises when one nuclide's abundance falls near or below the detection sensitivity threshold, where discriminative spectral features become indistinguishable from statistical fluctuations in the composite spectrum.

\begin{figure}[htbp]
    \centering
    \includegraphics[width=\textwidth]{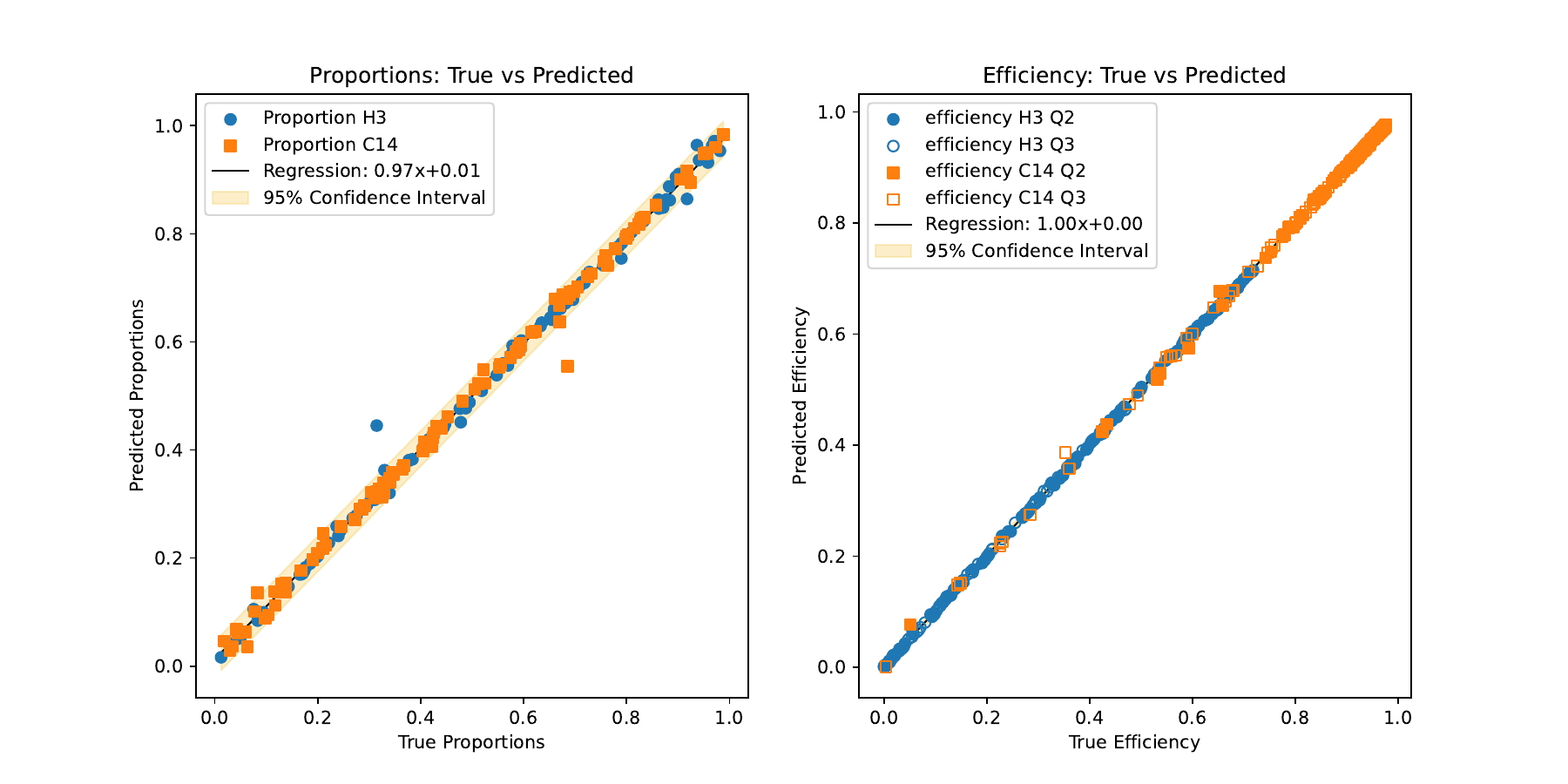}
    \caption{True-predicted value comparisons for multi-task TDCR model outputs. Left: Activity proportion predictions for $^3$H (blue dots) and $^{14}$C (orange squares) against true values. The regression line and 95\% confidence interval illustrate high correlation. Right: Detection efficiency predictions for double-tube and triple-tube coincidence configurations (Q2/Q3) of $^3$H and $^{14}$C. The near-ideal regression line and tight confidence interval confirm the model’s precision across quenching levels. These results demonstrate strong model agreement with true values for both activity and efficiency.}
    \label{fig:AandQ}
\end{figure}

In efficiency estimation for double-tube and triple-tube coincidence configurations, the MSE ($1 \times 10^{-5}$) and MAE ($0.002$) reflected exceptional accuracy, with scatter plots confirming strong linear correlation ($R^2 > 0.99$) across all quenching levels . This precision validates the model's capacity to learn complex relationships between spectral features and detection efficiency without relying on manual correction curves. Performance  declined for both activity and efficiency for severely quenched cases (efficiency $< 0.01$) due to sparse discriminative features. Strong quenching compresses spectral information into fewer data points, limiting the model's ability to discern nuclide-specific characteristics.  The performance is also shown visually in the  Fig.\ref{fig:AandQ}, both activity and efficiency predictions clustered tightly around the 1:1 diagonal regression lines with slope close to one. In general, the current model yields more accurate predictions for efficiency than for activity, which can be further fine-tuned by adjusting the loss weights between tasks.

\begin{figure}[htbp]
    \centering
    \includegraphics[width=\textwidth]{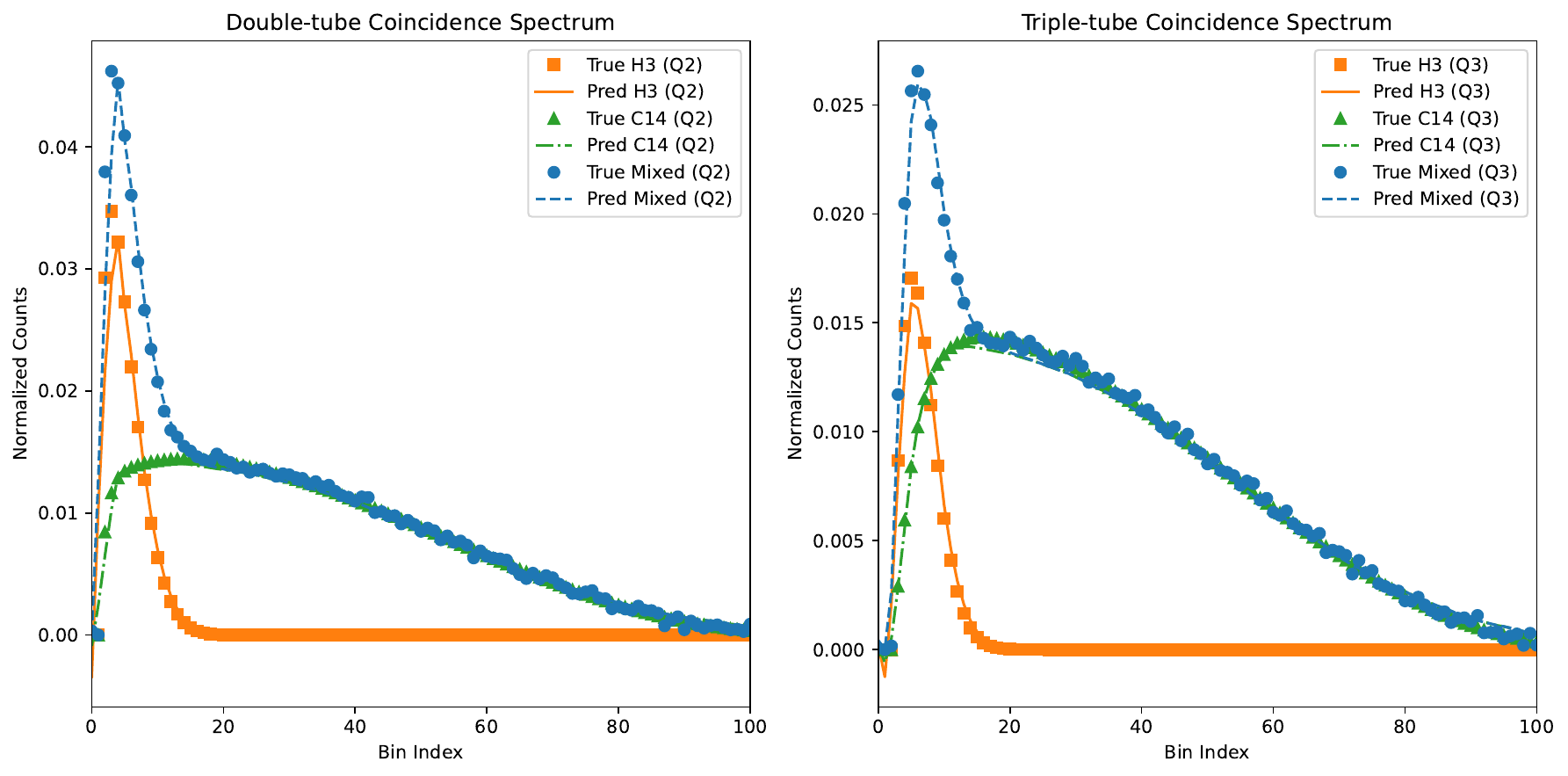} 
    \caption{Normalized count distributions for double-tube (left) and triple-tube (right) coincidence spectra. Curves compare true (H3, C14, mixed) and predicted (symbols/lines) spectra, validating the model’s ability to reconstruct TDCR spectral shapes for individual nuclides and mixtures.}
    \label{fig:spectra}
\end{figure}

Spectral reconstruction quality was rigorously quantified by the SSIM of predicted and truth spectra pairs. The SSIM of $0.9998$ indicates near-identical perceptual quality between reconstructed and truth spectra. Figure~\ref{fig:spectra} shows reconstructed spectra together with ground truth spectra demonstrating the model's ability to preserve  distribution features across the entire energy range. The two-stage training protocol prioritized accurate activity and efficiency predictions by deferring spectral reconstruction to Stage 2 of joint fine-tuning. This staging prevented the larger spectral datasets from dominating parameter updates during initial training, ensuring fundamental quantitative tasks received preferential optimization before full-spectrum learning.

The foundation of the model relies on the fact that the integration of double-tube and triple-tube coincidence data provides complementary information about nuclide-specific light yield responses, thereby enhancing robustness to quenching effect prediction. The model's success stems from its multi-task architecture that utilizes shared quenching induced spectral features across tasks, enhancing overall learning efficiency. Unlike traditional TDCR methods that rely on manual quenching correction curves, this framework eliminates human bias by learning relationships directly from data and enables simultaneous estimation of proportions, efficiencies, and spectra, reducing processing time. The deployed model achieves operational readiness within seconds of loading, enabling near-instantaneous sample predictions.

Current limitations include reduced accuracy for extreme quenching  and pronounced activity asymmetries between components, primarily attributable to  inadequate feature separability in parameter space. Notably, architectural modifications such as convolutional neural networks by treating Q2/Q3 spectra as 2D inputs, attention mechanisms to highlight discriminative bins, and residual blocks to enhance gradient flow were also explored, but none yielded statistically significant improvements in key metrics of MSE and SSIM. This lack of improvement likely stems from the intrinsic characteristics of TDCR spectral data. Unlike natural images where CNNs excel at capturing local spatial hierarchies, Q2/Q3 spectra are inherently 1D signals with globally correlated features while quenching affects the entire spectral shape uniformly, and nuclide specific patterns span broad energy ranges rather than localized ``edges" or ``textures." Consequently, CNNs designed to exploit 2D spatial locality failed to extract meaningful additional information from the stacked Q2/Q3 input format. Similarly, attention mechanisms, which thrive on emphasizing sparse critical features in high-dimensional data, found limited utility here: TDCR spectra lack isolated ``important" bins, as even low-count regions contribute to efficiency calibration and proportion estimation. Residual blocks, while beneficial for very deep networks ($>50$ layers), offered no advantage in our relatively shallow architecture ($\sim$10–15 layers), where gradient flow remained stable without skip connections. These findings underscore that architectural complexity must align with data properties. Our baseline multi-task framework, with its focus on shared 1D feature extraction and task-specific refinement, already captures the characteristic distributions of nuclide beta spectra and differential energy shifting effects of quenching on Q2/Q3 signals for TDCR systems. This suggests that further performance gains may require not just architectural tweaks but deeper integration of physical constraints. Future investigations will prioritize (1) simulation to real-world transfer validation via physics guided transfer learning, explicitly encoding non-linearity and time variant \cite{transfer2021}; (2) expanded nuclide coverage, for example, $^{33}$P and $^{35}$S); (3) advanced noise models, such as PMT dark counts, for severely quenched samples; and (4) Bayesian techniques to quantify the uncertainty in model predictions \cite{barberBRML2012}.

\section{Summary}
This study develops a multi-task neural network to automate spectral deconvolution of TDCR beta spectra to enable simultaneous quantification of activity concentrations and detection efficiencies for individual radionuclides in multi-component radioactive samples. This model eliminates manual spectral interpretation, bypasses convergence instabilities in traditional numerical solvers, and obviates the need for standard sources, thereby reducing user expertise requirements while maintaining analytical precision to enable real-time analysis with inherent algorithmic robustness. Deployment potential spans critical domains including environmental radio-surveillance and clinical radiopharmaceutical quality control, where rapid, high-fidelity nuclide-specific quantification is essential.
    
\section{Acknowledgments}
We would like to express our gratitude to Prof. Cheng Li for initiating insightful discussions on this topic which laid the foundation for the conceptual development of this work. This work is supported by the National Natural Science Foundation of China (NSFC) under Grant No. 14-547.
\bibliographystyle{elsarticle-num}
\bibliography{ref_0815}

\end{document}